# A Deep Features Based Approach Using Modified ResNet50 and Gradient Boosting for Visual Sentiments Classification

Muhammad Arslan
Department of Computer Science
Loyola University Chicago
Chicago, USA
marslan@luc.edu

Muhammad Mubeen
Department of Computer Science
University Of the People
California, USA
muhammad.mubeen@uopeople.edu

Arslan Akram
Facultty of Computer Science and
Information Technology
The Superior University Lahore
Lahore, Pakistan
aaadoula11@gmail.com

Saadullah Farooq Abbasi
Department of Electronic, Electrical
and Systems Engineering
University of Birmingham
Birmingham, U. K.
s.f.abbasi@bham.ac.uk

Muhammad Salman Ali
Facultty of Computer Science and
Information Technology
The Superior University Lahore
Lahore, Pakistan
msalmanali0683@gmail.com

Muhammad Usman Tariq
Facultty of Computer Science and
Information Technology
The Superior University Lahore
Lahore, Pakistan
tariqusman263@gmail.com

*Abstract*— The versatile nature of Visual Sentiment Analysis (VSA) is one reason for its rising profile. It isn't easy to efficiently manage social media data with visual information since previous research has concentrated on Sentiment Analysis (SA) of single modalities, like textual. In addition, most visual sentiment studies need to adequately classify sentiment because they are mainly focused on simply merging modal attributes without investigating their intricate relationships. This prompted the suggestion of developing a fusion of deep learning and machine learning algorithms. In this research, a deep feature-based method for multiclass classification has been used to extract deep features from modified ResNet50. Furthermore, gradient boosting algorithm has been used to classify photos containing emotional content. The approach is thoroughly evaluated on two benchmarked datasets, CrowdFlower and GAPED. Finally, cutting-edge deep learning and machine learning models were used to compare the proposed strategy. When compared to state-of-the-art approaches, the proposed method demonstrates exceptional performance on the datasets presented.

## I. INTRODUCTION

Prior sentiment analysis focused on textual data and made tremendous progress. Researchers are now studying multimodal emotion prediction, focusing on visual content emotional analysis to improve user experience with events or themes. The popularity of photos on Twitter, Facebook, Google Plus, Flickr, Instagram, and Snapchat shows their relevance in online expression [1]. South African multimedia and data research requires techniques to handle different, unsupervised scenarios as social media material grows [2]. Flickr and Instagram users share photos of their happiness, grief, and boredom, proving that photos may better express emotions than words. Visual material is essential to expressing emotions. Thus, text-based emotion analysis must improve. Online media allows unintended emotional analysis of photographs and videos, a solid social indicator in the information age.

The deep learning structure architecture works in unbounded computer vision. CNNs are deep visual recognition structures. CNN models are approaching human sight recognition [3]. The multi-layered mode uses a layer-smart transformation to learn feature representation from pixels. Large-scale training data and supervised algorithm learning help CNN models. Evaluating a picture's emotional tone requires creativity and work. Visual sentiment models detect a picture's polarity. Pre-trained deep-learning classification models improve visual sentiment analysis with unbalanced datasets [4]. ML models are inferior to transfer learning models for classification [5]. Gradient boosting was used to train Resnet 50 to predict and categorize photo sentiments using Crowdflower sentiment polarity and GAPED datasets [6]. The main contributions of the proposed study are highlighted below.

1. Using deep features from ResNet-50 in sentiment categorization highlights the reliability of sophisticated neural network representations.
2. The study improves sentiment prediction performance by combining deep learning feature extraction and classical machine learning approaches with XGBoost as the classification algorithm. The importance of gradient boosting is clear.
3. The suggested method outperforms state-of-the-art algorithms in sentiment analysis when deep features are integrated with gradient boosting, indicating its promise for real-world classification tasks.

Remainder of the proposed paper is arranged as: section 2 provides a literature review of recent methods for visual sentiment analysis. Materials and methods used for this research are described in section 3. Section 4 provides research findings and discussions. Finally, in last section conclusions and future work is discussed.

## II. LITERATURE REVIEW

In this section, a brief survey of the literature on picture sentiment prediction has been provided. Machine learning, deep learning models, and low-level and semantic information have been utilized for the sentiment analysis of photos. An image's emotional tone was predicted using pixel-level features and low-level properties such as texture and color [7]. S. Siersdorfer et al. used latent semantic analysis (LSA) and the Bag-of-Visual-Words (BoVW) method to categorize picture emotions [8]. They took on the challenge of detecting emotions in different visual domains. Using pictures of art or beauty as a basis for visual sentiment analysis, T. Rao et al. [9] achieved far better results than low-level features.

Using a BiLSTM-CNN architecture and multi-level feature fusion, T. Sun et al. [10] achieved 60.86% accuracy with emotion ROIs in photographs for sentiment classification. However, they encountered computational inefficiencies as a result of high-dimensional features. Despite a small sample size limiting model resilience, G. Chandrasekaran's CrowdFlower-based deep-learning models achieved an encouraging 75.00% accuracy [11]. A further study demonstrated the necessity for architectural enhancements and improved feature extraction by achieving 67.75% accuracy with ResNet and MlDrNet on the Emotion 6 dataset [12]. Using deep feature fusion, I. S. Al-Tameemi et al. presented multi-view sentiment analysis (MVSA) with an accuracy of 79.06% and suggested using bigger datasets for improved performance [13].

While S. Agarwal and M. K. Gupta achieved an accuracy of 73.80% for EMOTIC emotions classification using EfficientNet-7, they did suggest some ways to improve model performance [14]. Hyperparameter adjustment and additional data augmentation were investigated. Applying SVM with LBP to the FI Dataset, N. Desai et al. [15] achieved an accuracy of 76.01% while recommending tweaks to the model's parameters and feature extraction processes. Although feature fusion has the potential to improve performance, Karuthakannan and G. Velusamy only investigated neural networks and hybrid features for ROI emotion classification, which resulted in an accuracy rate of 80.75 percent [16]. The necessity to rectify data imbalance was brought to light by Jiang et al.'s 94.00% accuracy in classifying Twitter sentiments using VGG16 and SVM [17]. With a success rate of 74.00% and suggestions for more advanced feature extraction and model augmentation techniques, Fouad et al. [18] used DWT and Gabor Features with SVMs to categorize CrowdFlower feelings. This review showcases the successes and setbacks of visual sentiment analysis, highlighting the importance of resolving these issues in order to progress the discipline and showcase its potential in many domains.

## III. PROPOSED METHODOLOGY

The study trained ResNet-50, a top convolutional neural network, utilizing input photos because of its feature extraction capabilities. Future sentiment analysis uses these deep features to identify sophisticated emotional patterns. Gradient Boosting, a powerful ensemble learning technique, improved the sentiment categorization model's robustness and accuracy. This combination approach improves visual sentiment analysis by revealing image-evoked emotions.

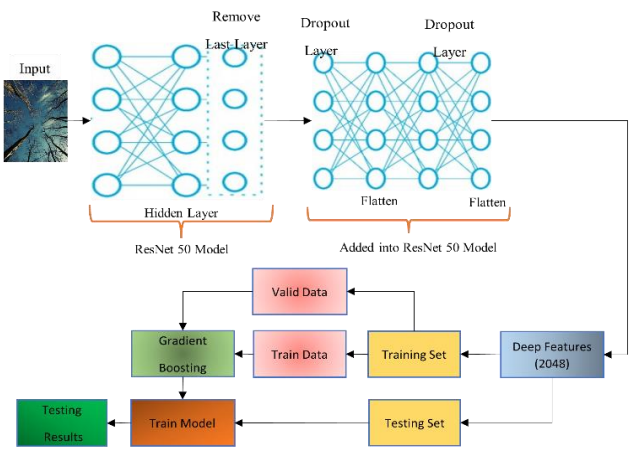

Fig. 1. Block diagram Visual sentiment analysis.

### A. Datasets

This study uses three benchmarked datasets to evaluate the suggested technique. The datasets are described below. The Geneva Affective Picture Database (GAPED) [19] provides visual emotion cues. Neutral photos were largely of inanimate items, whereas positive shots were of baby humans and animals in nature. Images were judged on emotional intensity, scene authenticity, and internal and outward morality. The dataset has several image URLs with their sentiment polarity annotated by annotators. It might be neutral, extremely negative, positive, or very positive. The proposed approach extracted 11,715 positive and negative pictures from this dataset [11].

TABLE I. DATASET DETAILS TO BE USED FOR EXPERIMENTS

| Dataset | Negative | Neutral | Positive | Total |
|---|---|---|---|---|
| GAPED | 520 | 90 | 122 | 732 |
| CrowdFlower | 1912 | 2023 | 7780 | 11715 |
| Total | 2432 | 2113 | 7902 | 12447 |

### B. Preprocessing

Scaling input images to the ResNet50 model's input size is essential to our image-based sentiment analysis approach because each image is unique. ResNet50 needs three RGB 224-by-224-pixel input images. The suggested method scales all input photos to this size before feeding them into the neural network for categorization to ensure model conformity. Scaling maintains input data structure and lets the model process photos.

### C. Deep ResNet50 Feature Extraction

One of the many benefits of deep feature extraction over more conventional approaches is that it does away with the requirement for human intervention when choosing kernel sizes and cropping image features. Human intervention is necessary for traditional feature extraction, which might only work with some picture formats. We intend to extract and merge deep features to improve picture-based emotion classification. After incorporating the final layers 'FC_1000,' 'FC1000_Softmax,' and 'FC_Classification,' the ResNet50 model was enhanced by incorporating 'FC_3,' 'FC3_Softmax,' and 'Class_output' layers to represent the 'Negative,' 'Neutral,' and 'Positive' classes, respectively. After the TL-ResNet50 model's three sentiment class layers, the 'avg_pool' layer contains 2048 neurons. Figure 2 shows the revised ResNet50 deep feature extraction in action, with blue blocks representing the block operations '3, 4, 6, and 3'. While blocks 'conv3_x,' 'conv4_x,' and 'conv5_x' each have 12, 18, and 9 layers, respectively, the 'conv2_x' block has nine layers due to three repeated layers. The residual function is responsible for ResNet's exceptional learning and generalization capabilities [20]. Deep learning helps the network automatically learn visual attributes layer by layer. For class output prediction, deep learning networks like ResNet50 use SoftMax classification in the last layer. Before FC layers, high-level properties can be retrieved. This study uses ResNet50 to extract high-level features from the fifth residual block (Conv5_x) in the 'avg_pool' layer. This output extracts deep-picture features. ResNet50 predicts 2048-dimensional deep feature vectors [21].

### D. Experimental Settup

Multiple tools and libraries were used for experimental analysis and model construction. Python [21], a flexible language with a huge library and good usability, was used to code the entire experiment. Machine learning models used XGBoost, a strong gradient-boosting framework. Gradient Boosting-based XGBOOST is considered. hyperparameters fine-tune XGBOOST classifier model performance. Regularization, n-estimators, Learning rate, and n-trees-depth are XGBOOST hyperparameters. In sequence, N-trees-depth, regularization, and learning rate fall into [0.05, 0.3], [max, min], and [lambda, alpha]. These parameters function best at 0.08, Max, and lambda. Optimizing these hyperparameters can help your XGBOOST classifier perform well and robustly in classification tasks.

## IV. RESULTS AND DISCUSSIONS

This section presents the findings of a large-scale experiment that tested the proposed strategy for breast cancer classification. To train and grade the models, the suggested method made use of the evaluation metrics specified in recent state-of-the-art methodologies. Information was collected from various performance evaluation instruments. The following domains were subject to the examinations:

These areas were the focus of the experiments:
1. Performance analysis of the proposed framework is measured for classification.
2. Performance analysis on the combined dataset, proposed method is used to evaluate the efficacy.
3. Comparative analysis of how the proposed method stacks up against recent methods.

### A. Evaluation of Proposed Method on CrowdFlower, GAPED, and Combined dataset

The method's efficacy and reliability are thoroughly assessed in this section. 10-fold cross-validation yields the best test set model testing results for each dataset. In Figure 3, the fold-wise comparison graph shows that the suggested method's accuracy is linear to almost all folds, making it applicable to any dataset. The graph compares our model's accuracy over ten cross-validation folds. A detailed analysis shows that most folds perform reliably with constant accuracy scores. However, small precision discrepancies between divisions imply that the data part may affect model performance. Our technique is robust and generalizable, with competitive accuracy across all folds. Good results suggest our algorithm can learn from data and forecast appropriately. In the future, we will identify performance variance sources and improve the model for consistency and reliability across datasets.

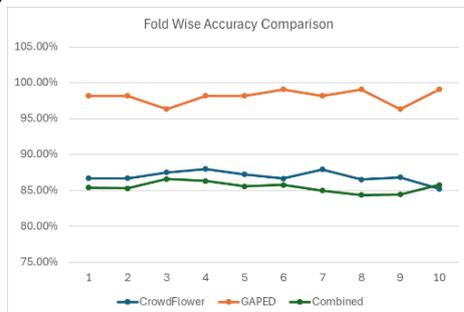

Fig. 2. Fold wise Accuracy comparison of proposed method on different datasets.

The ability to comprehend the underlying emotional tone of text data is crucial in areas such as social media analysis and customer service, and sentiment analysis provides one such tool. To improve the precision of sentiment classification, state-of-the-art machine learning methods such as XGBoost are utilized. Recent studies have shown that a proposed method may effectively classify emotions as positive, neutral, or negative with high recall (up to 90%), accuracy (98%), and F1 scores (83%-87%). This approach routinely achieves metrics that are outstanding and well-balanced, with AUC values above 89%. Its reliability and promise for real-world applications are proven.

TABLE II.   TESTING RESULTS OF BEST XGBOOST MODEL ON CROWDFLOWER, GAPED, AND COMBINED DATASETS

| Dataset | Class | Precision | Recall | F1-Score | AUC | Accuracy |
|---|---|---|---|---|---|---|
| Crowd Flower | Negative | 89% | 90% | 90% | 93% | **87%** |
| | Neutral | 90% | 84% | 87% | 90% | |
| | Positive | 82% | 87% | 85% | 89% | |
| GAPED | Negative | 97% | 97% | 97% | 98% | **98%** |
| | Neutral | 97% | 99% | 98% | 99% | |
| | Positive | 100% | 98% | 99% | 99% | |
| Combined | Negative | 87% | 87% | 87% | 90% | **86%** |
| | Neutral | 90% | 84% | 87% | 90% | |
| | Positive | 81% | 86% | 83% | 88% | |

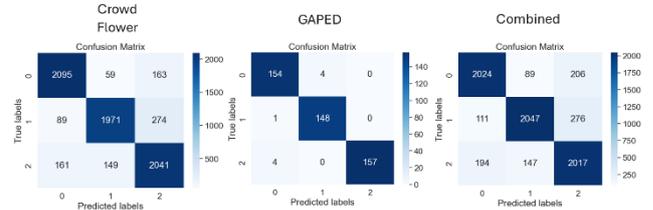

Fig. 3. Testing Confusion Matrix of Best XGBOOST Model on CrowdFlower, GAPED, Combined Dataset

The Figure 4. Receiver Operating Characteristic (ROC) curves, show the model's performance across Negative, Neutral, and Positive sentiment classes using optimal XGBOOST weights on CrowdFlower, GAPED, and combined datasets. Area Under the Curve (AUC) values for each sentiment class measure the model's ability to distinguish three datasets. The GAPED dataset has strong discriminating power with an AUC of 0.981. The merged dataset has strong sentiment discrimination with an AUC of 0.873. The CrowdFlower sentiment dataset ROC curve has an AUC of 0.812, indicating that it can distinguish positive emotions from others. The ROC curves and AUC values reveal that the XGBOOST model discriminates well across all sentiment datasets in this investigation.

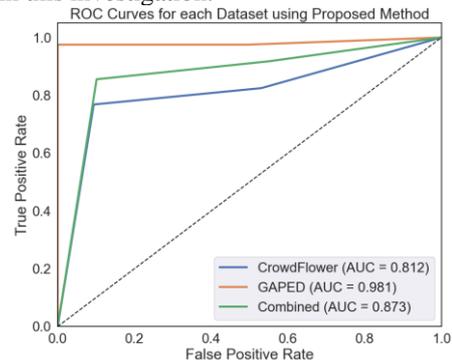

Fig. 4. Testing AUROC Graph of Best XGBOOST Model on CrowdFlower, GAPED, and Combined Dataset

### B. Comparison wih State of the art visual sentiment analysis methods.

Using Deep ResNet50 Features and Gradient Boosting, the suggested approach outperforms state-of-the-art methods, with an accuracy of 87% on the CrowdFlower dataset.

Conventional approaches could be more efficient and accurate by 60.86 to 76.01%. The proposed method demonstrates its resilience and adaptability with an accuracy of 98% on the GAPED dataset, even if deep learning models perform 73% to 73.80% better on datasets such as EMOTIC and CrowdFlower. On average, it gets 86% of the datasets right. Problems with data imbalance persist, though, and either more evenly distributed datasets or a balancing approach are required for the system to reach its full potential.

TABLE III.  COMPARATIVE ANALYSIS OF STATE OF THE ART METHOD WITH PROPOSED METHOD

| Ref, Year | Method | Dataset | Accuracy |
|---|---|---|---|
| [10], 2022 | BI-LSTM with Feature Fusion | EmotionROI | 60.86% |
| [11], 2022 | VGG19 | CrowdFlower | 73% |
| [11], 2022 | ResNet50V2 | CrowdFlower | 75% |
| [14], 2022 | EfficientNet-B7 | EMOTIC | 73.80% |
| [15], 2020 | CNN with Affective Regions | Self-Collected | 76.01% |
| [18], 2023 | Event Concept with Object Detection | CrowdFlower | 74% |
| **Modified ResNet50 Features with Gradient Boosting** | | **CrowdFlower** | **87%** |
| | | **GAPED** | **98%** |
| | | **Combined** | **86%** |

## V. CONCLUSION AND FUTURE WORK

Using ResNet50 to distinguish textures and gradient boosting to classify features, this research suggests a visual sentiment analysis approach. An all-encompassing framework for evaluating attitudes toward things is produced by this method, which optimizes interactions between visual modalities. With an accuracy of 86% on the combined dataset and 98% on GAPED, the approach considerably enhances sentiment classification. Improving the model's capacity to process massive datasets and expanding multimodal sentiment analysis to include audio and video will be the primary goals of future efforts. The approach can be made more widely used in many applications by making it more adaptable and scalable.